\documentclass[conference,compsoc]{IEEEtran}
\ifCLASSOPTIONcompsoc
  \usepackage[nocompress]{cite}
\else
  \usepackage{cite}
\fi
\ifCLASSINFOpdf
\else
\fi
\usepackage{tikz}
\usetikzlibrary{shapes,calendar,matrix,backgrounds,folding,arrows}
\usepackage{pgfplots}

\tikzstyle{redrect}=[rectangle, draw=black, fill=red, text centered, rounded corners]
\tikzstyle{greenrect} = [rectangle, draw=black, fill=green, text centered, rounded corners]

\tikzset{
decision/.style = {diamond, draw, fill=blue!20,
  text width=4.5em, text badly centered, node distance=3cm, inner sep=0pt},
block/.style = {rectangle, draw, fill=blue!20,
  text width=3em, text centered, rounded corners, minimum height=4em},
textdata/.style = {rectangle,
  text width=3em, text centered, minimum height=4em},
procedure/.style = {rectangle, draw, fill=blue!20,
  text width=3.5em, text centered,  minimum height=3em,  minimum width=3em},
line/.style = {draw,very thick,  -latex'},
line1/.style = {draw,thick, <->},
compose/.style = {rectangle, draw, fill=none, rounded corners, blue!60, thick},
compose1/.style = {rectangle, draw, fill=none, orange!60, thick},
compose3/.style = {rectangle, draw, fill=none, green!60, thick},
compose4/.style = {rectangle, draw, fill=none, red!60, thick},
compose5/.style = {rectangle, draw, fill=none, rounded corners, red, thick, dashed},
compose2/.style = {rectangle, draw, fill=none, black!60, thick},
cloud/.style = {draw, ellipse,fill=red!20, node distance=3cm,
  minimum height=2em},
subroutine/.style = {draw,rectangle split,
  rectangle split parts=3,minimum height=1cm,
  rectangle split part fill={red!50, green!50, blue!20, yellow!50}},
connector/.style = {draw,circle,node distance=3cm,fill=yellow!20},
data/.style = {draw, trapezium, text width=3em, align=center, fill=olive!20, trapezium left angle=75pt, trapezium right angle=105, trapezium stretches = true, minimum height=3em, minimum width = 3em}
}

\usepackage{amsmath}
\usepackage{amssymb}

\usepackage{array}

\ifCLASSOPTIONcompsoc
  \usepackage[caption=false,font=footnotesize,labelfont=sf,textfont=sf]{subfig}
\else
  \usepackage[caption=false,font=footnotesize]{subfig}
\fi
\begin{document}
%
\title{A Bi-LSTM-RNN Model for Relation Classification\\ Using Low-Cost Sequence Features}



%
\author{\IEEEauthorblockN{Fei Li$^1$,
Meishan Zhang$^2$,
Guohong Fu$^2$,
Tao Qian$^3$ and
Donghong Ji$^1$$^*$}
\IEEEauthorblockA{$^1$School of Computer, Wuhan University, Wuhan, China\\Email: \textbraceleft lifei\_csnlp, dhji\textbraceright @whu.edu.cn}
\IEEEauthorblockA{$^2$School of Computer Science and Technology, Heilongjiang University, China}
\IEEEauthorblockA{$^3$College of Computer Science and Technology, Hubei University of Science and Technology, China}
}


\maketitle

\begin{abstract}
Relation classification is associated with many potential applications in the artificial intelligence area. Recent approaches usually leverage neural networks based on structure features such as syntactic or dependency features to solve this problem. However, high-cost structure features make such approaches inconvenient to be directly used. In addition, structure features are probably domain-dependent. Therefore, this paper proposes a bi-directional long-short-term-memory recurrent-neural-network (Bi-LSTM-RNN) model based on low-cost sequence features to address relation classification. This model divides a sentence or text segment into five parts, namely two target entities and their three contexts. It learns the representations of entities and their contexts, and uses them to classify relations. We evaluate our model on two standard benchmark datasets in different domains, namely SemEval-2010 Task 8 and BioNLP-ST 2016 Task BB3. In the former dataset, our model achieves comparable performance compared with other models using sequence features. In the latter dataset, our model obtains the third best results compared with other models in the official evaluation. Moreover, we find that the context between two target entities plays the most important role in relation classification. Furthermore, statistic experiments show that the context between two target entities can be used as an approximate replacement of the shortest dependency path when dependency parsing is not used.

\end{abstract}


%
\IEEEpeerreviewmaketitle

\section{Introduction}
Relation classification is associated with many potential applications in the artificial intelligence area such as information extraction, question answering and semantic network construction. In the natural language processing (NLP) community, there are a number of evaluation tasks \cite{2004:lrec:george,hendrickx-EtAl:2010:SemEval,bossy-EtAl:2013:BioNLPST,OverviewBB2016} about relation classification. They aim to classify the relations between two target entities into some predefined relation types. For example, \textquotedblleft{}burst\textquotedblright{} and \textquotedblleft{}pressure\textquotedblright{} have a \textquotedblleft{}Cause-Effect\textquotedblright{} relation in the sentence \textquotedblleft{}The burst has been caused by water hammer pressure.\textquotedblright{}.

Early studies \cite{zhang-EtAl:2006:COLACL,chan-roth:2011:ACL-HLT2011,2014:entityrelation:li,bmc:2015:parisa} mainly focused on feature-based or kernel-based approaches to solve this problem, but they need to pay much attention on feature engineering or kernel design. Recently, the approaches based on deep neural networks such as convolutional neural networks (CNNs) \cite{zeng-EtAl:2014:Coling}, recursive neural networks (RecursiveNNs) \cite{socher-EtAl:2012:EMNLP-CoNLL} and recurrent neural networks (RNNs) \cite{xu-EtAl:2015:EMNLP2} have become increasingly popular in order to reduce manual intervention. In these approaches, structure features (e.g., syntactic or dependency features) are usually effective, since they can help models to remove less relevant noise and get more compact representations.

However, structure features may cause some problems: on the one hand, the high cost for parsing sentences makes such approaches inconvenient to be directly used; on the other hand, syntactic or dependency parsers are probably domain-dependent. For example, a parser trained in news corpora may be imprecise when it is used in biomedical text, which will unavoidably hurt the performance of models using structure features.

This paper proposes a Bi-LSTM-RNN model based on low-cost sequence features to address relation classification. Our motivation is that the relation between two target entities can be represented by the entities and contexts surrounding them. Therefore, the Bi-LSTM-RNN model firstly performs bi-directional recurrent computation along all the tokens of the sentences which the relation spans. Then, the sequence of token representations, which are generated in the previous step, is divided into five parts according to the order that tokens occur in these sentences:
\begin{itemize}
\item \emph{\textbf{before}} context, which consists of the tokens before the former target entity;
\item \emph{\textbf{former}} entity, which consists of the tokens in the former target entity;
\item \emph{\textbf{middle}} context, which consists of the tokens between two target entities;
\item \emph{\textbf{latter}} entity, which consists of the tokens in the latter target entity;
\item \emph{\textbf{after}} context, which consists of the tokens after the latter target entity.
\end{itemize}
Some relation examples are shown as below.
\begin{itemize}
\item Message-Topic: $[$In this comprehensive$]$$_{before}$ $[$guide$]$$_{former}$ $[$, over 850$]$$_{middle}$ $[$roses$]$$_{latter}$ $[$are described, illustrated, and arranged by group.$]$$_{after}$
\item Lives-In: $[$ $]$$_{before}$ $[$Vibrio salmonicida$]$$_{former}$ $[$was detected in sediment samples from diseased farms. It was also detected in a$]$$_{middle}$ $[$sediment sample from a disease-free fish farm$]$$_{latter}$ $[$.$]$$_{after}$
\end{itemize}
After the sequence of token representations has been divided, standard pooling functions are applied over the token representations of each part, and we obtain five representations corresponding to the five parts. Lastly, they are concatenated and fed into a softmax layer for relation classification. To avoid the need of structure features, our model uses low-cost sequence features such as words and part-of-speech (POS) tags. Moreover, LSTMs \cite{lstm:1998:hochr} are used to attenuate the gradient vanishing problem when two target entities are distant in text.

We evaluate our model on two standard benchmark datasets in different domains, namely SemEval-2010 Task 8 \cite{hendrickx-EtAl:2010:SemEval} and BioNLP-ST 2016 Task BB3 \cite{OverviewBB2016}. Experimental results in the former dataset show that our model achieves comparable performance compared with other models that use sequence features. In the latter dataset, our model obtains the third best results compared with other models in the official evaluation. In addition, we evaluate the contributions of three contexts, and find that the \emph{middle} context plays the most important role in relation classification. Furthermore, statistic experiments show that the \emph{middle} context can be used as an approximate replacement of the shortest dependency path when dependency parsing is not used. Our model is implemented using LibN3L \cite{lrec:2016:libn3l}, and the code is publicly available under GPL at: http://xxxxx.

\section{Related Work}
Early approaches for relation classification are usually feature/kernel-based. Feature-based approaches \cite{chan-roth:2011:ACL-HLT2011,2014:entityrelation:li} design a great number of lexical, syntactic or semantic features and use classifiers such as support vector machines (SVMs) to classify relations. The problem may be that handcrafted features are labor-consuming and time-costing. Kernel-based approaches \cite{zhang-EtAl:2006:COLACL,plank-moschitti:2013:ACL2013} do not need much effort on feature engineering, but well-designed kernel functions, which are usually based on syntactic or dependency structures, are crucial for relation classification.

Recently, the approaches based on deep neural networks become new research hotspots for relation classification, since they can achieve promising results with less manual intervention. RecursiveNNs \cite{socher-EtAl:2012:EMNLP-CoNLL,ebrahimi-dou:2015:NAACL-HLT} are firstly used for this task to learn sentence representations along syntactic or dependency structures. Liu et al. \cite{liu-EtAl:2015:ACL-IJCNLP} combine RecursiveNNs and CNNs to capture features of the shortest dependency path and its attached subtree. Zeng et al. \cite{zeng-EtAl:2014:Coling} leverage CNNs to classify relations with lexical, sentence and word position features. Based on CNNs, dos Santos et al. \cite{dossantos-xiang-zhou:2015:ACL-IJCNLP} propose a novel ranking loss function for special treatment of the noisy \emph{Other} class. Xu et al. \cite{xu-EtAl:2015:EMNLP1} leverage CNNs to learn representations from shortest dependency paths, and address the relation directionality by special treatment on sampling. Yu et al. \cite{nips:2014:yumo} propose a factor-based embedding model to decompose sentences into factors based on linguistic annotations, extract features and combine them via sum-pooling. Xu et al. \cite{xu-EtAl:2015:EMNLP2} use multi-channel RNNs along the shortest dependency path between two target entities, and they obtain the best result without any special treatment. Most of the approaches above use structure features. In this paper, we follow the line of RNNs, but not use structure features.

Since some classical work was published \cite{icml:2007:hiton,jmlr:2011:cw}, deep neural networks have received increasing research attention in the NLP community. They have been successfully applied into many other NLP tasks, such as sentiment analysis \cite{2015:oosenti:zhang,2016:aaai:zhang}, parsing \cite{2014:nndep:chen,zhou-EtAl:2015:ACL-IJCNLP3} and machine translation \cite{cho-EtAl:2014:EMNLP2014,luong-pham-manning:2015:EMNLP}. To tackle different problems, prior work used various networks such as CNNs \cite{LeCun:1998:convol} or RNNs \cite{mikilov:2010:lanmodel}, and some optimization technologies \cite{2011:adagrad:duchi}. Recently, some researchers turn their attention to new unsupervised learning technologies and the ability of deep models to generalize well from small datasets \cite{2015:nature:lecunbengiohinton}. However, non-neural approaches are still important and attract considerable research attention, since neural networks seem not to outperform other approaches in all the tasks.

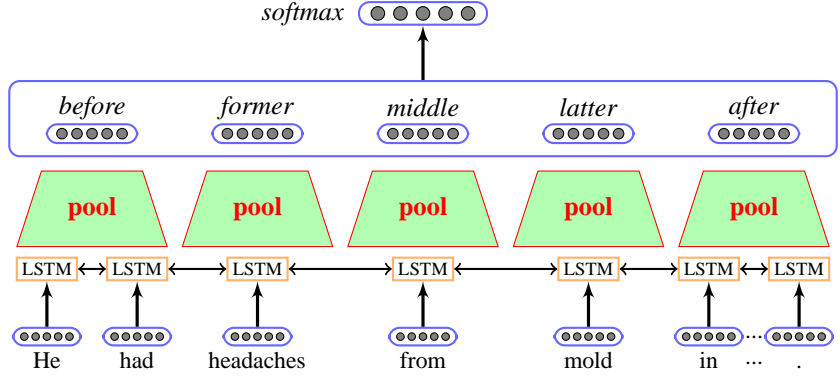
\begin{figure*}[t]
\begin{center}

\begin{tikzpicture}[node distance = 2.0, auto]

\node [circle,draw,minimum size=1.8mm, inner sep=0pt, fill=gray] (falsenode) {};
\node [circle,draw,minimum size=1.8mm, inner sep=0pt, fill=gray, left of=falsenode, node distance = 0.3cm] (truenode) {};
\node [circle,draw,minimum size=1.8mm, inner sep=0pt, fill=gray, right of=falsenode, node distance = 0.3cm] (neutralnode) {};
\node [circle,draw,minimum size=1.8mm, inner sep=0pt, fill=gray, left of=truenode, node distance = 0.3cm] (truenode1) {};
\node [circle,draw,minimum size=1.8mm, inner sep=0pt, fill=gray, right of=neutralnode, node distance = 0.3cm] (neutralnode1) {};
\node [compose, below of = falsenode, minimum height = 0.3cm, minimum width = 1.7cm, node distance = 0.0cm]  (output){};

\node [inner sep=0pt, left of=falsenode, node distance = 1.6cm, font=\normalsize] (typename) {\emph{softmax}};

\node [circle,draw,minimum size=1.4mm, inner sep=0pt, fill=gray, below of=falsenode, node distance = 1.6cm] (h1Tmidnode) {};
\node [circle,draw,minimum size=1.4mm, inner sep=0pt, fill=gray, left of=h1Tmidnode, node distance = 0.4cm] (h1Tleftnode1) {};
\node [circle,draw,minimum size=1.4mm, inner sep=0pt, fill=gray, right of=h1Tmidnode, node distance = 0.4cm] (h1Trightnode1) {};
\node [circle,draw,minimum size=1.4mm, inner sep=0pt, fill=gray, left of=h1Tmidnode, node distance = 0.2cm] (h1Tleftnode2) {};
\node [circle,draw,minimum size=1.4mm, inner sep=0pt, fill=gray, right of=h1Tmidnode, node distance = 0.2cm] (h1Trightnode2) {};
\node [compose, below of = h1Tmidnode, minimum height = 0.25cm, minimum width = 1.2cm, node distance = 0.0cm]  (h1Tpool){};

\node [circle,draw,minimum size=1.4mm, inner sep=0pt, fill=gray, left of=h1Tmidnode, node distance = 2.2cm] (h1Lmidnode) {};
\node [circle,draw,minimum size=1.4mm, inner sep=0pt, fill=gray, left of=h1Lmidnode, node distance = 0.4cm] (h1Lleftnode1) {};
\node [circle,draw,minimum size=1.4mm, inner sep=0pt, fill=gray, right of=h1Lmidnode, node distance = 0.4cm] (h1Lrightnode1) {};
\node [circle,draw,minimum size=1.4mm, inner sep=0pt, fill=gray, left of=h1Lmidnode, node distance = 0.2cm] (h1Lleftnode2) {};
\node [circle,draw,minimum size=1.4mm, inner sep=0pt, fill=gray, right of=h1Lmidnode, node distance = 0.2cm] (h1Lrightnode2) {};
\node [compose, below of = h1Lmidnode, minimum height = 0.25cm, minimum width = 1.2cm, node distance = 0.0cm]  (h1Lpool){};

\node [circle,draw,minimum size=1.4mm, inner sep=0pt, fill=gray, left of=h1Lmidnode, node distance = 2.2cm] (h1LLmidnode) {};
\node [circle,draw,minimum size=1.4mm, inner sep=0pt, fill=gray, left of=h1LLmidnode, node distance = 0.4cm] (h1LLleftnode1) {};
\node [circle,draw,minimum size=1.4mm, inner sep=0pt, fill=gray, right of=h1LLmidnode, node distance = 0.4cm] (h1LLrightnode1) {};
\node [circle,draw,minimum size=1.4mm, inner sep=0pt, fill=gray, left of=h1LLmidnode, node distance = 0.2cm] (h1LLleftnode2) {};
\node [circle,draw,minimum size=1.4mm, inner sep=0pt, fill=gray, right of=h1LLmidnode, node distance = 0.2cm] (h1LLrightnode2) {};
\node [compose, below of = h1LLmidnode, minimum height = 0.25cm, minimum width = 1.2cm, node distance = 0.0cm]  (h1LLpool){};

\node [circle,draw,minimum size=1.4mm, inner sep=0pt, fill=gray, right of=h1Tmidnode, node distance = 2.2cm] (h1Rmidnode) {};
\node [circle,draw,minimum size=1.4mm, inner sep=0pt, fill=gray, left of=h1Rmidnode, node distance = 0.4cm] (h1Rleftnode1) {};
\node [circle,draw,minimum size=1.4mm, inner sep=0pt, fill=gray, right of=h1Rmidnode, node distance = 0.4cm] (h1Rrightnode1) {};
\node [circle,draw,minimum size=1.4mm, inner sep=0pt, fill=gray, left of=h1Rmidnode, node distance = 0.2cm] (h1Rleftnode2) {};
\node [circle,draw,minimum size=1.4mm, inner sep=0pt, fill=gray, right of=h1Rmidnode, node distance = 0.2cm] (h1Rrightnode2) {};
\node [compose, below of = h1Rmidnode, minimum height = 0.25cm, minimum width = 1.2cm, node distance = 0.0cm]  (h1Rpool){};

\node [circle,draw,minimum size=1.4mm, inner sep=0pt, fill=gray, right of=h1Rmidnode, node distance = 2.2cm] (h1RRmidnode) {};
\node [circle,draw,minimum size=1.4mm, inner sep=0pt, fill=gray, left of=h1RRmidnode, node distance = 0.4cm] (h1RRleftnode1) {};
\node [circle,draw,minimum size=1.4mm, inner sep=0pt, fill=gray, right of=h1RRmidnode, node distance = 0.4cm] (h1RRrightnode1) {};
\node [circle,draw,minimum size=1.4mm, inner sep=0pt, fill=gray, left of=h1RRmidnode, node distance = 0.2cm] (h1RRleftnode2) {};
\node [circle,draw,minimum size=1.4mm, inner sep=0pt, fill=gray, right of=h1RRmidnode, node distance = 0.2cm] (h1RRrightnode2) {};
\node [compose, below of = h1RRmidnode, minimum height = 0.25cm, minimum width = 1.2cm, node distance = 0.0cm]  (h1RRpool){};

\node [inner sep=0pt, minimum size=0, above of=h1Tmidnode, node distance = 0.35cm, font=\normalsize] (targettext) {\emph{middle}};
\node [inner sep=0pt, minimum size=0, above of=h1Lmidnode, node distance = 0.35cm, font=\normalsize] (lefttext) {\emph{former}};
\node [inner sep=0pt, minimum size=0, above of=h1Rmidnode, node distance = 0.35cm, font=\normalsize] (righttext) {\emph{latter}};
\node [inner sep=0pt, minimum size=0, above of=h1LLmidnode, node distance = 0.35cm, font=\normalsize] (llefttext) {\emph{before}};
\node [inner sep=0pt, minimum size=0, above of=h1RRmidnode, node distance = 0.35cm, font=\normalsize] (rrighttext) {\emph{after}};

\node [compose, above of = h1Tmidnode, minimum height = 1cm, minimum width = 11cm, node distance = 0.2cm]  (h1pool){};
\path [line] (h1pool) -- (output);

\node [trapezium, trapezium angle=72,draw, red, fill=green!30, below of=h1Lmidnode, node distance = 1cm, minimum width=2cm] () {\strut \bf pool};

\node [trapezium, trapezium angle=72,draw, red, fill=green!30, below of=h1Tmidnode, node distance = 1cm, minimum width=2cm] () {\strut \bf pool};

\node [trapezium, trapezium angle=72,draw, red, fill=green!30, below of=h1Rmidnode, node distance = 1cm, minimum width=2cm] () {\strut \bf pool};

\node [trapezium, trapezium angle=72,draw, red, fill=green!30, below of=h1LLmidnode, node distance = 1cm, minimum width=2.0cm] () {\strut \bf pool};

\node [trapezium, trapezium angle=72,draw, red, fill=green!30, below of=h1RRmidnode, node distance = 1cm, minimum width=2cm] () {\strut \bf pool};

\node [inner sep=0pt, below of=h1Lmidnode, node distance = 1.8cm, font=\scriptsize] (h2Lp0) {};

\node [minimum size=1.0mm, inner sep=0pt, left of=h2Lp0, node distance = 0cm, font=\scriptsize] (h2Lp1m) {LSTM};
\node [compose1, below of = h2Lp1m, minimum height = 0.3cm, minimum width = 0.8cm, node distance = 0.0cm]  (h2Lp1comp){};

\node [inner sep=0pt, below of=h1Tmidnode, node distance = 1.8cm, font=\scriptsize] (h2Tp0) {};

\node [minimum size=1.0mm, inner sep=0pt, left of=h2Tp0, node distance = 0cm, font=\scriptsize] (h2Tp1m) {LSTM};
\node [compose1, below of = h2Tp1m, minimum height = 0.3cm, minimum width = 0.8cm, node distance = 0.0cm]  (h2Tp1comp){};

\node [inner sep=0pt, below of=h1Rmidnode, node distance = 1.8cm, font=\scriptsize] (h2Rp0) {};

\node [minimum size=1.0mm, inner sep=0pt, left of=h2Rp0, node distance = 0cm, font=\scriptsize] (h2Rp1m) {LSTM};
\node [compose1, below of = h2Rp1m, minimum height = 0.3cm, minimum width = 0.8cm, node distance = -0.0cm]  (h2Rp1comp){};

\node [inner sep=0pt, below of=h1LLmidnode, node distance = 1.8cm, font=\scriptsize] (h2LLp0) {};

\node [minimum size=1.0mm, inner sep=0pt, left of=h2LLp0, node distance = 0.6cm, font=\scriptsize] (h2LLp1m) {LSTM};
\node [compose1, below of = h2LLp1m, minimum height = 0.3cm, minimum width = 0.8cm, node distance = 0.0cm]  (h2LLp1comp){};

\node [minimum size=1.0mm, inner sep=0pt, right of=h2LLp0, node distance = 0.6cm, font=\scriptsize] (h2LLp2m) {LSTM};
\node [compose1, below of = h2LLp2m, minimum height = 0.3cm, minimum width = 0.8cm, node distance = 0.0cm]  (h2LLp2comp){};

\path [line1] (h2LLp1comp) -- (h2LLp2comp);

\node [inner sep=0pt, below of=h1RRmidnode, node distance = 1.8cm, font=\scriptsize] (h2RRp0) {};

\node [minimum size=1.0mm, inner sep=0pt, left of=h2RRp0, node distance = 0.6cm, font=\scriptsize] (h2RRp1m) {LSTM};
\node [compose1, below of = h2RRp1m, minimum height = 0.3cm, minimum width = 0.8cm, node distance = -0.0cm]  (h2RRp1comp){};

\node [minimum size=1.0mm, inner sep=0pt, right of=h2RRp0, node distance = 0.6cm, font=\scriptsize] (h2RRp2lm) {LSTM};
\node [compose1, below of = h2RRp2lm, minimum height = 0.3cm, minimum width = 0.8cm, node distance = -0.0cm]  (h2RRp2comp){};

\path [line1] (h2RRp1comp) -- (h2RRp2comp);

\path [line1] (h2Lp1comp) -- (h2Tp1comp);
\path [line1] (h2Tp1comp) -- (h2Rp1comp);
\path [line1] (h2LLp2comp) -- (h2Lp1comp);
\path [line1] (h2Rp1comp) -- (h2RRp1comp);

\node [inner sep=0pt, below of=h2Lp0, node distance = 0.9cm, font=\scriptsize] (h3Lp0) {};
\node [inner sep=0pt, below of=h3Lp0, node distance = 0.35cm, font=\scriptsize] (h3Lp0Text) {};

\node [circle,draw,minimum size=1.0mm, inner sep=0pt, fill=gray, left of=h3Lp0, node distance = 0cm] (h3Lp1m) {};
\node [circle,draw,minimum size=1.0mm, inner sep=0pt, fill=gray, left of=h3Lp1m, node distance = 0.15cm] (h3Lp1l) {};
\node [circle,draw,minimum size=1.0mm, inner sep=0pt, fill=gray, right of=h3Lp1m, node distance = 0.15cm] (h3Lp1r) {};
\node [circle,draw,minimum size=1.0mm, inner sep=0pt, fill=gray, left of=h3Lp1l, node distance = 0.15cm] (h3Lp1ll) {};
\node [circle,draw,minimum size=1.0mm, inner sep=0pt, fill=gray, right of=h3Lp1r, node distance = 0.15cm] (h3Lp1rr) {};
\node [compose, below of = h3Lp1m, minimum height = 0.2cm, minimum width = 0.9cm, node distance = 0.0cm]  (h3Lp1comp){};
\node [inner sep=0pt, below of=h3Lp1m, node distance = 0.35cm, font=\small] (h3Lp1text) {\strut  headaches};

\path [line] (h3Lp1comp) -- (h2Lp1comp);

\node [inner sep=0pt, below of=h2Tp0, node distance = 0.9cm, font=\scriptsize] (h3Tp0) {};
\node [inner sep=0pt, below of=h3Tp0, node distance = 0.35cm, font=\scriptsize] (h3Tp0Text) {};

\node [circle,draw,minimum size=1.0mm, inner sep=0pt, fill=gray, left of=h3Tp0, node distance = 0cm] (h3Tp1m) {};
\node [circle,draw,minimum size=1.0mm, inner sep=0pt, fill=gray, left of=h3Tp1m, node distance = 0.15cm] (h3Tp1l) {};
\node [circle,draw,minimum size=1.0mm, inner sep=0pt, fill=gray, right of=h3Tp1m, node distance = 0.15cm] (h3Tp1r) {};
\node [circle,draw,minimum size=1.0mm, inner sep=0pt, fill=gray, left of=h3Tp1l, node distance = 0.15cm] (h3Tp1ll) {};
\node [circle,draw,minimum size=1.0mm, inner sep=0pt, fill=gray, right of=h3Tp1r, node distance = 0.15cm] (h3Tp1rr) {};
\node [compose, below of = h3Tp1m, minimum height = 0.2cm, minimum width = 0.9cm, node distance = 0.0cm]  (h3Tp1comp){};
\node [inner sep=0pt, below of=h3Tp1m, node distance = 0.35cm, font=\small] (h3Tp1text) {\strut  from};

\path [line] (h3Tp1comp) -- (h2Tp1comp);

\node [inner sep=0pt, below of=h2Rp0, node distance = 0.9cm, font=\scriptsize] (h3Rp0) {};
\node [inner sep=0pt, below of=h3Rp0, node distance = 0.35cm, font=\scriptsize] (h3Rp0Text) {};

\node [circle,draw,minimum size=1.0mm, inner sep=0pt, fill=gray, left of=h3Rp0, node distance = 0cm] (h3Rp1m) {};
\node [circle,draw,minimum size=1.0mm, inner sep=0pt, fill=gray, left of=h3Rp1m, node distance = 0.15cm] (h3Rp1l) {};
\node [circle,draw,minimum size=1.0mm, inner sep=0pt, fill=gray, right of=h3Rp1m, node distance = 0.15cm] (h3Rp1r) {};
\node [circle,draw,minimum size=1.0mm, inner sep=0pt, fill=gray, left of=h3Rp1l, node distance = 0.15cm] (h3Rp1ll) {};
\node [circle,draw,minimum size=1.0mm, inner sep=0pt, fill=gray, right of=h3Rp1r, node distance = 0.15cm] (h3Rp1rr) {};
\node [compose, below of = h3Rp1m, minimum height = 0.2cm, minimum width = 0.9cm, node distance = -0.0cm]  (h3Rp1comp){};
\node [inner sep=0pt, below of=h3Rp1m, node distance = 0.35cm, font=\small] (h3Rp1text) {\strut mold};

\path [line] (h3Rp1comp) -- (h2Rp1comp);

\node [inner sep=0pt, below of=h2LLp0, node distance = 0.9cm, font=\scriptsize] (h3LLp0) {};
\node [inner sep=0pt, below of=h3LLp0, node distance = 0.35cm, font=\scriptsize] (h3LLp0Text) {};

\node [circle,draw,minimum size=1.0mm, inner sep=0pt, fill=gray, left of=h3LLp0, node distance = 0.6cm] (h3LLp1m) {};
\node [circle,draw,minimum size=1.0mm, inner sep=0pt, fill=gray, left of=h3LLp1m, node distance = 0.15cm] (h3LLp1l) {};
\node [circle,draw,minimum size=1.0mm, inner sep=0pt, fill=gray, right of=h3LLp1m, node distance = 0.15cm] (h3LLp1r) {};
\node [circle,draw,minimum size=1.0mm, inner sep=0pt, fill=gray, left of=h3LLp1l, node distance = 0.15cm] (h3LLp1ll) {};
\node [circle,draw,minimum size=1.0mm, inner sep=0pt, fill=gray, right of=h3LLp1r, node distance = 0.15cm] (h3LLp1rr) {};
\node [compose, below of = h3LLp1m, minimum height = 0.2cm, minimum width = 0.9cm, node distance = 0.0cm]  (h3LLp1comp){};
\node [inner sep=0pt, below of=h3LLp1m, node distance = 0.35cm, font=\small] (h3LLp1text) {\strut  He};

\path [line] (h3LLp1comp) -- (h2LLp1comp);

\node [circle,draw,minimum size=1.0mm, inner sep=0pt, fill=gray, right of=h3LLp0, node distance = 0.6cm] (h3LLp2m) {};
\node [circle,draw,minimum size=1.0mm, inner sep=0pt, fill=gray, left of=h3LLp2m, node distance = 0.15cm] (h3LLp2l) {};
\node [circle,draw,minimum size=1.0mm, inner sep=0pt, fill=gray, right of=h3LLp2m, node distance = 0.15cm] (h3LLp2r) {};
\node [circle,draw,minimum size=1.0mm, inner sep=0pt, fill=gray, left of=h3LLp2l, node distance = 0.15cm] (h3LLp2ll) {};
\node [circle,draw,minimum size=1.0mm, inner sep=0pt, fill=gray, right of=h3LLp2r, node distance = 0.15cm] (h3LLp2rr) {};
\node [compose, below of = h3LLp2m, minimum height = 0.2cm, minimum width = 0.9cm, node distance = 0.0cm]  (h3LLp2comp){};
\node [inner sep=0pt, below of=h3LLp2m, node distance = 0.35cm, font=\small] (h3LLp2text) {\strut had};

\path [line] (h3LLp2comp) -- (h2LLp2comp);

\node [inner sep=0pt, below of=h2RRp0, node distance = 0.9cm, font=\small] (h3RRp0) {...};
\node [inner sep=0pt, below of=h3RRp0, node distance = 0.35cm, font=\small] (h3RRp0Text) {...};

\node [circle,draw,minimum size=1.0mm, inner sep=0pt, fill=gray, left of=h3RRp0, node distance = 0.6cm] (h3RRp1m) {};
\node [circle,draw,minimum size=1.0mm, inner sep=0pt, fill=gray, left of=h3RRp1m, node distance = 0.15cm] (h3RRp1l) {};
\node [circle,draw,minimum size=1.0mm, inner sep=0pt, fill=gray, right of=h3RRp1m, node distance = 0.15cm] (h3RRp1r) {};
\node [circle,draw,minimum size=1.0mm, inner sep=0pt, fill=gray, left of=h3RRp1l, node distance = 0.15cm] (h3RRp1ll) {};
\node [circle,draw,minimum size=1.0mm, inner sep=0pt, fill=gray, right of=h3RRp1r, node distance = 0.15cm] (h3RRp1rr) {};
\node [compose, below of = h3RRp1m, minimum height = 0.2cm, minimum width = 0.9cm, node distance = -0.0cm]  (h3RRp1comp){};
\node [inner sep=0pt, below of=h3RRp1m, node distance = 0.35cm, font=\small] (h3RRp1text) {\strut in};

\path [line] (h3RRp1comp) -- (h2RRp1comp);

\node [circle,draw,minimum size=1.0mm, inner sep=0pt, fill=gray, right of=h3RRp0, node distance = 0.6cm] (h3RRp2m) {};
\node [circle,draw,minimum size=1.0mm, inner sep=0pt, fill=gray, left of=h3RRp2m, node distance = 0.15cm] (h3RRp2l) {};
\node [circle,draw,minimum size=1.0mm, inner sep=0pt, fill=gray, right of=h3RRp2m, node distance = 0.15cm] (h3RRp2r) {};
\node [circle,draw,minimum size=1.0mm, inner sep=0pt, fill=gray, left of=h3RRp2l, node distance = 0.15cm] (h3RRp2ll) {};
\node [circle,draw,minimum size=1.0mm, inner sep=0pt, fill=gray, right of=h3RRp2r, node distance = 0.15cm] (h3RRp2rr) {};
\node [compose, below of = h3RRp2m, minimum height = 0.2cm, minimum width = 0.9cm, node distance = -0.0cm]  (h3RRp2comp){};
\node [inner sep=0pt, below of=h3RRp2m, node distance = 0.35cm, font=\small] (h3RRp2text) {\strut .};

\path [line] (h3RRp2comp) -- (h2RRp2comp);

\end{tikzpicture}

\caption{An illustration of the Bi-LSTM-RNN model. The example is \textquotedblleft{}He had \emph{headaches}$_{e_1}$ from \emph{mold}$_{e_2}$ in the bedrooms.\textquotedblright{}. \emph{e}$_1$ and \emph{e}$_2$ denote two target entities.
} \label{fig:lstmrnn}
\end{center}
\vspace{-0.3cm}
\end{figure*}

\section{Our Bi-LSTM-RNN Model}
Our model has several characters: relation classification is modeled based on entity and context representations learned from LSTM-RNNs; only low-cost sequence features are used to avoid the problems of structure features; features are extracted from bi-directional RNNs using simple pooling technologies; relations between entities that occur in different sentences can also be classified.

\subsection{Long Short Term Memory (LSTM)}
LSTMs \cite{lstm:1998:hochr} aim to facilitate the training of RNNs by solving the diminishing and exploding gradient problems in the deep or long structures. It can be defined as below: given an input sequence \emph{\textbf{x}} = \{\emph{x}$_1$, \emph{x}$_2$, ..., \emph{x}$_n$\}, LSTMs associate each of them with an input gate (\emph{i}$_t$), a forget gate (\emph{f}$_t$), an output gate (\emph{o}$_t$), a candidate cell state (\emph{\~c}$_{t}$), a cell state (\emph{c}$_{t}$) and a hidden state (\emph{h}$_{t}$). \emph{i}$_{t}$ decides what new information will be stored in the current cell state \emph{c}$_{t}$. \emph{f}$_t$ decides what information is going to be thrown away from the previous cell state \emph{c}$_{t-1}$. \emph{o}$_{t}$ decides what information will be output to the current hidden state \emph{h}$_{t}$ (\emph{n}$^{(lstm)}$ dimension), which is computed by
\begin{equation} \label{eqn:lstm}
\begin{split}
\emph{i}_{t}\ &=\ \sigma(\ \emph{W}^{(i)}\ \cdot\ (\ \emph{h}_{t-1}\ \oplus\ \emph{x}_{t}\ )\ +\ b^{(i)}), \\
\emph{f}_{t}\ &=\ \sigma(\ \emph{W}^{(f)}\ \cdot\ (\ \emph{h}_{t-1}\ \oplus\ \emph{x}_{t}\ )\ +\ b^{(f)}), \\
\emph{o}_{t}\ &=\ \sigma(\ \emph{W}^{(o)}\ \cdot\ (\ \emph{h}_{t-1}\ \oplus\ \emph{x}_{t}\ )\ +\ b^{(o)}), \\
\emph{\~c}_{t} \ &=\ tanh(\ \emph{W}^{(c)}\ \cdot\ (\ \emph{h}_{t-1}\ \oplus\ \emph{x}_{t}\ )\ +\ b^{(c)}), \\
\emph{c}_{t}\ &=\ \emph{f}_{t}\ \times\ \emph{c}_{t-1}\ +\ \emph{i}_{t}\ \times\ \emph{\~c}_{t},\\
\emph{h}_t\ &=\ \emph{o}_{t}\ \times\ tanh(\ \emph{c}_{t}\ ),
\end{split}
\end{equation}
where \emph{$\sigma$} denotes the sigmoid function. + and $\times$ denote the element-wise addition and product operations, respectively. $\oplus$ denotes the vector concatenation. The input, forget, output gate and candidate cell state are associated with their own weight matrices \emph{W} and bias vectors \emph{b}, which are learned.

\subsection{Bi-LSTM-RNN}

The framework of our Bi-LSTM-RNN model is shown in Figure \ref{fig:lstmrnn}. The given sentence or text segment can be considered as a token sequence \emph{\textbf{s}} = \{\emph{s}$_1$, \emph{s}$_2$, ..., \emph{s}$_n$\}. A LSTM unit takes the embedding \emph{x}$_t$ of each token \emph{s}$_t$ as input and outputs a hidden state \emph{h}$'_t$ computed by Equation \ref{eqn:lstm}. Then we will get a hidden state sequence \emph{\textbf{h}}$'$ = \{\emph{h}$'_1$, \emph{h}$'_2$, ..., \emph{h}$'_n$\} after the LSTM unit has finished recurrent computation along all the tokens from left to right. Here \emph{h}$'_t$ does not only capture the information of token \emph{s}$_t$, but also that of its predecessors. To capture the information of its successors, a counterpart \emph{h}$''_t$ of \emph{h}$'_t$ is also generated by another LSTM unit computing in the reverse direction.

The final representation sequence of all the tokens, namely \emph{\textbf{h}} = \{\emph{h}$_1$, \emph{h}$_2$, ..., \emph{h}$_n$\}, is generated by concatenating \emph{h}$'_t$ and \emph{h}$''_t$ at first, and then using a compositional operation to reduce the dimension to \emph{n}$^{(h)}$. This procedure can be formulated as
\begin{equation} \label{eqn:project}
\emph{h}_t\ =\ tanh(\ W_1\ \cdot\ (\ \emph{h}'_t\ \oplus\ \emph{h}''_t\ )\ +\ b_1).
\end{equation}

In the following step, we divide the token representation sequence \emph{\textbf{h}} into five parts, namely \emph{before}, \emph{former}, \emph{middle}, \emph{latter} and \emph{after} according to the boundaries of target entities. Four standard pooling functions (i.e., \emph{max}, \emph{min}, \emph{avg}, \emph{std}) are respectively applied over the token representations of each part and we obtain five representations corresponding to the five parts. For example, the \emph{former} entity representation r$_{former}$ can be computed by
\begin{equation} \label{eqn:pool}
\begin{split}
&r_{max_j}\ =\ \max_{1 \leqslant k \leqslant K}\ h_{k_j},\\
&r_{min_j}\ =\ \min_{1 \leqslant k \leqslant K}\ h_{k_j},\\
&r_{avg_j}\ =\ \frac{1}{K}\sum_{1 \leqslant k \leqslant K}h_{k_j},\\
&r_{std_j}\ =\ \sqrt{\sum_{1 \leqslant k \leqslant K}h_{k_j}^2},
\end{split}
\end{equation}
\begin{equation} \label{eqn:pool:concat}
r_{former}\ =\ r_{max}\ \oplus\ r_{min}\ \oplus\ r_{avg}\ \oplus\ r_{std},
\end{equation}
where the \emph{former} entity is assumed to start at the 1st token and end at the \emph{K}-th token. \emph{h}$_{k_j}$ denotes the \emph{j}-th component of the \emph{k}-th token representation vector. \emph{r}$_{max_j}$, \emph{r}$_{min_j}$, \emph{r}$_{avg_j}$ and \emph{r}$_{std_j}$ denote the \emph{j}-th components of representation vectors generated by the corresponding pooling functions.

The penultimate layer of our Bi-LSTM-RNN model consists of the concatenation of five representations corresponding to entities and their contexts, which can be formulated by
\begin{equation} \label{eqn:penul}
\emph{x}_{penul} = \emph{r}_{before} \oplus \emph{r}_{former} \oplus \emph{r}_{middle} \oplus \emph{r}_{latter} \oplus \emph{r}_{after}.
\end{equation}

Finally, the output layer calculates the probabilities of all relation types, so that the one with the maximum probability is selected. The probability of the \emph{i}-th relation type \emph{R}$_i$ is computed by
\begin{equation}
p(\ R_i\ ) = softmax(\ R_i\ )\ =\ \frac{e^{\emph{w}_{2_{i}}\ \cdot\ \emph{x}_{penul}}}{\sum_{j=1}^{|R|} e^{\emph{w}_{2_{j}}\ \cdot\ \emph{x}_{penul}}},
\label{eqn:softmax}
\end{equation}
where \emph{w}$_{2_{i}}$ denotes the \emph{i}-th row of parameter matrix \emph{W}${_{2}}$ in the output layer.

\subsection{Training}
Given a set of annotated training examples, the training objective of our model is to minimize the cross-entropy loss, with a L$_2$ regularization term, given by
\begin{equation} \label{eqn:object}
L(\ \theta\ )\ =\ -\sum_{i} \log{p_{g_{i}}}\ +\ \frac{\beta}{2}\ \lVert\ \theta\ \rVert^2_2,
\end{equation}
where $\theta$ denotes all the parameters of the model. \emph{p}$_{g_{i}}$ indicates the probability of the gold relation type of the \emph{i}-th training example as given by the model. $\beta$ is the regularization parameter.

We employ standard training frameworks for the model, namely stochastic gradient decent using AdaGrad \cite{2011:adagrad:duchi}. Derivatives are calculated from standard back-propagation \cite{1996:backprop:goller}. More details will be further described in Section \ref{sec:experiment}.

\subsection{Features}
\begin{figure}[t]
\centering

\subfloat[][Feature Usage]{
\centering
\begin{tikzpicture}[node distance = 2.0, auto]

\node [minimum size=1.0mm, inner sep=0pt, font=\scriptsize] (lstm) {LSTM};
\node [compose1, below of = lstm, minimum height = 0.3cm, minimum width = 0.9cm, node distance = 0.0cm]  (lstmcomp){};

\node [minimum size=1.0mm, inner sep=0, below of=lstm, node distance = 0.8cm, font=\scriptsize] (concat) {$\oplus$};
\node [compose2, below of = concat, minimum height = 0.3cm, minimum width = 0.8cm, node distance = 0.0cm]  (concatcomp){};

\path [line] (concatcomp) -- (lstmcomp)node [midway] (m2){\emph{x}};

\node [circle,draw,minimum size=1.0mm, inner sep=0pt, fill=gray, below of=concat, node distance = 0.8cm, xshift=0cm] (charMid) {};
\node [circle,draw,minimum size=1.0mm, inner sep=0pt, fill=gray, left of=charMid, node distance = 0.15cm] (charLeft) {};
\node [circle,draw,minimum size=1.0mm, inner sep=0pt, fill=gray, right of=charMid, node distance = 0.15cm] (charRight) {};
\node [circle,draw,minimum size=1.0mm, inner sep=0pt, fill=gray, left of=charLeft, node distance = 0.15cm] (charLeftl) {};
\node [circle,draw,minimum size=1.0mm, inner sep=0pt, fill=gray, right of=charRight, node distance = 0.15cm] (charRightr) {};
\node [compose, below of = charMid, minimum height = 0.2cm, minimum width = 0.9cm, node distance = -0.0cm]  (charComp){};
\node [inner sep=0pt, below of=charMid, node distance = 0.35cm, font=\small] (char) {\strut \emph{r}$_{char}$};

\node [circle,draw,minimum size=1.0mm, inner sep=0pt, fill=gray, below of=concat, node distance = 0.8cm, xshift=-1cm] (wordMid) {};
\node [circle,draw,minimum size=1.0mm, inner sep=0pt, fill=gray, left of=wordMid, node distance = 0.15cm] (wordLeft) {};
\node [circle,draw,minimum size=1.0mm, inner sep=0pt, fill=gray, right of=wordMid, node distance = 0.15cm] (wordRight) {};
\node [circle,draw,minimum size=1.0mm, inner sep=0pt, fill=gray, left of=wordLeft, node distance = 0.15cm] (wordLeftl) {};
\node [circle,draw,minimum size=1.0mm, inner sep=0pt, fill=gray, right of=wordRight, node distance = 0.15cm] (wordRightr) {};
\node [compose, below of = wordMid, minimum height = 0.2cm, minimum width = 0.9cm, node distance = -0.0cm]  (wordComp){};
\node [inner sep=0pt, below of=wordMid, node distance = 0.35cm, font=\small] (word) {\strut \emph{r}$_{ran}$};

\node [circle,draw,minimum size=1.0mm, inner sep=0pt, fill=gray, below of=concat, node distance = 0.8cm, xshift=-2cm] (prewordMid) {};
\node [circle,draw,minimum size=1.0mm, inner sep=0pt, fill=gray, left of=prewordMid, node distance = 0.15cm] (prewordLeft) {};
\node [circle,draw,minimum size=1.0mm, inner sep=0pt, fill=gray, right of=prewordMid, node distance = 0.15cm] (prewordRight) {};
\node [circle,draw,minimum size=1.0mm, inner sep=0pt, fill=gray, left of=prewordLeft, node distance = 0.15cm] (prewordLeftl) {};
\node [circle,draw,minimum size=1.0mm, inner sep=0pt, fill=gray, right of=prewordRight, node distance = 0.15cm] (prewordRightr) {};
\node [compose, below of = prewordMid, minimum height = 0.2cm, minimum width = 0.9cm, node distance = -0.0cm]  (prewordComp){};
\node [inner sep=0pt, below of=prewordMid, node distance = 0.35cm, font=\small] (preword) {\strut \emph{r}$_{pre}$};

\node [circle,draw,minimum size=1.0mm, inner sep=0pt, fill=gray, below of=concat, node distance = 0.8cm, xshift=1cm] (posMid) {};
\node [circle,draw,minimum size=1.0mm, inner sep=0pt, fill=gray, left of=posMid, node distance = 0.15cm] (posLeft) {};
\node [circle,draw,minimum size=1.0mm, inner sep=0pt, fill=gray, right of=posMid, node distance = 0.15cm] (posRight) {};
\node [circle,draw,minimum size=1.0mm, inner sep=0pt, fill=gray, left of=posLeft, node distance = 0.15cm] (posLeftl) {};
\node [circle,draw,minimum size=1.0mm, inner sep=0pt, fill=gray, right of=posRight, node distance = 0.15cm] (posRightr) {};
\node [compose, below of = posMid, minimum height = 0.2cm, minimum width = 0.9cm, node distance = -0.0cm]  (posComp){};
\node [inner sep=0pt, below of=posMid, node distance = 0.35cm, font=\small] (pos) {\strut \emph{r}$_{pos}$};

\node [circle,draw,minimum size=1.0mm, inner sep=0pt, fill=gray, below of=concat, node distance = 0.8cm, xshift=2cm] (wnMid) {};
\node [circle,draw,minimum size=1.0mm, inner sep=0pt, fill=gray, left of=wnMid, node distance = 0.15cm] (wnLeft) {};
\node [circle,draw,minimum size=1.0mm, inner sep=0pt, fill=gray, right of=wnMid, node distance = 0.15cm] (wnRight) {};
\node [circle,draw,minimum size=1.0mm, inner sep=0pt, fill=gray, left of=wnLeft, node distance = 0.15cm] (wnLeftl) {};
\node [circle,draw,minimum size=1.0mm, inner sep=0pt, fill=gray, right of=wnRight, node distance = 0.15cm] (wnRightr) {};
\node [compose, below of = wnMid, minimum height = 0.2cm, minimum width = 0.9cm, node distance = -0.0cm]  (wnComp){};
\node [inner sep=0pt, below of=wnMid, node distance = 0.35cm, font=\small] (wnh) {\strut \emph{r}$_{wnh}$};

\path [line] (prewordComp) -- (concatcomp);
\path [line] (wordComp) -- (concatcomp);
\path [line] (charComp) -- (concatcomp);
\path [line] (posComp) -- (concatcomp);
\path [line] (wnComp) -- (concatcomp);

\node [minimum size=1.0mm, inner sep=0, below of=char, node distance = 0.8cm, font=\scriptsize] (echar) {Fig. 2b};
\node [compose2, below of = echar, minimum height = 0.4cm, minimum width = 1cm, node distance = 0.0cm]  (echarcomp){};

\node [minimum size=1.0mm, inner sep=0, below of=word, node distance = 0.8cm, font=\scriptsize] (eword) {\emph{E}$_{ran}$};
\node [compose2, below of = eword, minimum height = 0.4cm, minimum width = 0.8cm, node distance = 0.0cm]  (ewordcomp){};

\node [minimum size=1.0mm, inner sep=0, below of=preword, node distance = 0.8cm, font=\scriptsize] (epreword) {\emph{E}$_{pre}$};
\node [compose2, below of = epreword, minimum height = 0.4cm, minimum width = 0.8cm, node distance = 0.0cm]  (eprewordcomp){};

\node [minimum size=1.0mm, inner sep=0, below of=pos, node distance = 0.8cm, font=\scriptsize] (epos) {\emph{E}$_{pos}$};
\node [compose2, below of = epos, minimum height = 0.4cm, minimum width = 0.8cm, node distance = 0.0cm]  (eposcomp){};

\node [minimum size=1.0mm, inner sep=0, below of=wnh, node distance = 0.8cm, font=\scriptsize] (ewnh) {\emph{E}$_{wnh}$};
\node [compose2, below of = ewnh, minimum height = 0.4cm, minimum width = 0.8cm, node distance = 0.0cm]  (ewnhcomp){};

\path [line] (echarcomp) -- (char);
\path [line] (ewordcomp) -- (word);
\path [line] (eprewordcomp) -- (preword);
\path [line] (eposcomp) -- (pos);
\path [line] (ewnhcomp) -- (wnh);

\node [inner sep=0pt, below of=echarcomp, node distance = 0.5cm, font=\small] () {\strut d,o,g};
\node [inner sep=0pt, below of=ewordcomp, node distance = 0.5cm, font=\small] () {\strut dog};
\node [inner sep=0pt, below of=eprewordcomp, node distance = 0.5cm, font=\small] () {\strut dog};
\node [inner sep=0pt, below of=eposcomp, node distance = 0.5cm, font=\small] () {\strut NN};
\node [inner sep=0pt, below of=ewnhcomp, node distance = 0.5cm, font=\small] () {\strut animal};

\end{tikzpicture}
\label{fig:features:concat}
}

\hfil

\subfloat[][Character Feature Usage]{
\centering
\begin{tikzpicture}[node distance = 2.0, auto]

\node [inner sep=0pt, minimum size=0, font=\small] (r) {\strut \emph{r}$_{char}$};

\node [circle,draw,minimum size=1.4mm, inner sep=0pt, fill=gray, below of=r, node distance = 0.3cm] (h1Tmidnode) {};
\node [circle,draw,minimum size=1.4mm, inner sep=0pt, fill=gray, left of=h1Tmidnode, node distance = 0.4cm] (h1Tleftnode1) {};
\node [circle,draw,minimum size=1.4mm, inner sep=0pt, fill=gray, right of=h1Tmidnode, node distance = 0.4cm] (h1Trightnode1) {};
\node [circle,draw,minimum size=1.4mm, inner sep=0pt, fill=gray, left of=h1Tmidnode, node distance = 0.2cm] (h1Tleftnode2) {};
\node [circle,draw,minimum size=1.4mm, inner sep=0pt, fill=gray, right of=h1Tmidnode, node distance = 0.2cm] (h1Trightnode2) {};
\node [compose, below of = h1Tmidnode, minimum height = 0.25cm, minimum width = 1.2cm, node distance = 0.0cm]  (h1Tpool){};

\node [minimum size=1.0mm, inner sep=0, below of=h1Tmidnode, node distance = 0.8cm, font=\scriptsize] (concat) {$\oplus$};
\node [compose2, below of = concat, minimum height = 0.3cm, minimum width = 0.8cm, node distance = 0.0cm]  (concatcomp){};

\path [line] (concatcomp) -- (h1Tpool);

\node [minimum size=1.0mm, inner sep=0pt, below of=concat, node distance = 1.2cm, font=\scriptsize] (lstm1) {LSTM};
\node [compose4, below of = lstm1, minimum height = 0.3cm, minimum width = 0.8cm, node distance = 0.0cm]  (lstm1comp){};

\node [minimum size=1.0mm, inner sep=0pt, left of=lstm1, node distance = 1.2cm, font=\scriptsize] (lstm2) {LSTM};
\node [compose4, below of = lstm2, minimum height = 0.3cm, minimum width = 0.8cm, node distance = 0.0cm]  (lstm2comp){};

\node [minimum size=1.0mm, inner sep=0pt, right of=lstm1, node distance = 1.2cm, font=\scriptsize] (lstm3) {LSTM};
\node [compose4, below of = lstm3, minimum height = 0.3cm, minimum width = 0.8cm, node distance = 0.0cm]  (lstm3comp){};

\draw [thick,->,red!60] (lstm2comp)--(lstm1comp);
\draw [thick,->,red!60] (lstm1comp)-- (lstm3comp);
\draw [thick,->,red!60,rounded corners=15pt] (lstm3comp.east) -- ++(0.5,0.5)-- (concatcomp.east) node [midway] (m1){};
\node [minimum size=1.0mm, inner sep=0pt, right of=m1, node distance = 0.8cm, font=\small, yshift=0.2cm] (m1text) {\emph{l2r}};

\node [minimum size=1.0mm, inner sep=0pt, below of=concat, node distance = 0.6cm, font=\scriptsize, xshift=-0.5cm] (lstm1r) {LSTM};
\node [compose3, below of = lstm1r, minimum height = 0.3cm, minimum width = 0.8cm, node distance = 0.0cm]  (lstm1rcomp){};

\node [minimum size=1.0mm, inner sep=0pt, left of=lstm1r, node distance = 1.2cm, font=\scriptsize] (lstm2r) {LSTM};
\node [compose3, below of = lstm2r, minimum height = 0.3cm, minimum width = 0.8cm, node distance = 0.0cm]  (lstm2rcomp){};

\node [minimum size=1.0mm, inner sep=0pt, right of=lstm1r, node distance = 1.2cm, font=\scriptsize] (lstm3r) {LSTM};
\node [compose3, below of = lstm3r, minimum height = 0.3cm, minimum width = 0.8cm, node distance = 0.0cm]  (lstm3rcomp){};

\draw [thick,->,green!60] (lstm3rcomp)--(lstm1rcomp);
\draw [thick,->,green!60] (lstm1rcomp)-- (lstm2rcomp);
\draw [thick,->,green!60,rounded corners=15pt] (lstm2rcomp.west) -- ++(-0.4,0.4)-- (concatcomp.west)node [midway] (m2){};
\node [minimum size=1.0mm, inner sep=0pt, left of=m2, node distance = 0.2cm, font=\small, yshift=0.1cm] (m2text) {\emph{r2l}};

\node [circle,draw,minimum size=1.0mm, inner sep=0pt, fill=gray, below of=lstm1comp, node distance = 0.8cm] (e1) {};
\node [circle,draw,minimum size=1.0mm, inner sep=0pt, fill=gray, left of=e1, node distance = 0.15cm] (e1l) {};
\node [circle,draw,minimum size=1.0mm, inner sep=0pt, fill=gray, right of=e1, node distance = 0.15cm] (e1r) {};
\node [circle,draw,minimum size=1.0mm, inner sep=0pt, fill=gray, left of=e1l, node distance = 0.15cm] (e1ll) {};
\node [circle,draw,minimum size=1.0mm, inner sep=0pt, fill=gray, right of=e1r, node distance = 0.15cm] (e1rr) {};
\node [compose, below of = e1, minimum height = 0.2cm, minimum width = 0.9cm, node distance = 0.0cm]  (e1comp){};
\node [minimum size=1.0mm, inner sep=0, below of=e1, node distance = 0.8cm, font=\scriptsize] (e1embmatrix) {\emph{E}$_{char}$};
\node [compose2, below of = e1embmatrix, minimum height = 0.4cm, minimum width = 0.8cm, node distance = 0.0cm]  (e1embmatrixcomp){};
\node [inner sep=0pt, below of=e1embmatrixcomp, node distance = 0.5cm, font=\small] (e1text) {\strut  o};
\path [line](e1embmatrixcomp) -- (e1comp);
\draw [thick,->](e1comp) -- (lstm1comp);
\draw [thick,->,rounded corners=15pt] (e1comp.west) -- ++(-0.2,0.4)-- (lstm1rcomp);

\node [circle,draw,minimum size=1.0mm, inner sep=0pt, fill=gray, below of=lstm2comp, node distance = 0.8cm] (e2) {};
\node [circle,draw,minimum size=1.0mm, inner sep=0pt, fill=gray, left of=e2, node distance = 0.15cm] (e2l) {};
\node [circle,draw,minimum size=1.0mm, inner sep=0pt, fill=gray, right of=e2, node distance = 0.15cm] (e2r) {};
\node [circle,draw,minimum size=1.0mm, inner sep=0pt, fill=gray, left of=e2l, node distance = 0.15cm] (e2ll) {};
\node [circle,draw,minimum size=1.0mm, inner sep=0pt, fill=gray, right of=e2r, node distance = 0.15cm] (e2rr) {};
\node [compose, below of = e2, minimum height = 0.2cm, minimum width = 0.9cm, node distance = 0.0cm]  (e2comp){};
\node [minimum size=1.0mm, inner sep=0, below of=e2, node distance = 0.8cm, font=\scriptsize] (e2embmatrix) {\emph{E}$_{char}$};
\node [compose2, below of = e2embmatrix, minimum height = 0.4cm, minimum width = 0.8cm, node distance = 0.0cm]  (e2embmatrixcomp){};
\node [inner sep=0pt, below of=e2embmatrixcomp, node distance = 0.5cm, font=\small] (e2text) {\strut  d};
\path [line](e2embmatrixcomp) -- (e2comp);
\draw [thick,->](e2comp) -- (lstm2comp);
\draw [thick,->,rounded corners=15pt] (e2comp.west) -- ++(-0.2,0.4)-- (lstm2rcomp);

\node [circle,draw,minimum size=1.0mm, inner sep=0pt, fill=gray, below of=lstm3comp, node distance = 0.8cm] (e3) {};
\node [circle,draw,minimum size=1.0mm, inner sep=0pt, fill=gray, left of=e3, node distance = 0.15cm] (e3l) {};
\node [circle,draw,minimum size=1.0mm, inner sep=0pt, fill=gray, right of=e3, node distance = 0.15cm] (e3r) {};
\node [circle,draw,minimum size=1.0mm, inner sep=0pt, fill=gray, left of=e3l, node distance = 0.15cm] (e3ll) {};
\node [circle,draw,minimum size=1.0mm, inner sep=0pt, fill=gray, right of=e3r, node distance = 0.15cm] (e3rr) {};
\node [compose, below of = e3, minimum height = 0.2cm, minimum width = 0.9cm, node distance = 0.0cm]  (e3comp){};
\node [minimum size=1.0mm, inner sep=0, below of=e3, node distance = 0.8cm, font=\scriptsize] (e3embmatrix) {\emph{E}$_{char}$};
\node [compose2, below of = e3embmatrix, minimum height = 0.4cm, minimum width = 0.8cm, node distance = 0.0cm]  (e3embmatrixcomp){};
\node [inner sep=0pt, below of=e3embmatrixcomp, node distance = 0.5cm, font=\small] (e3text) {\strut  g};
\path [line](e3embmatrixcomp) -- (e3comp);
\draw [thick,->] (e3comp) -- (lstm3comp);
\draw [thick,->,rounded corners=15pt] (e3comp.west) -- ++(-0.2,0.4)-- (lstm3rcomp);

\end{tikzpicture}
\label{fig:features:char}
}

\caption{Feature usage in Bi-LSTM-RNN.
} \label{fig:features}
\end{figure}
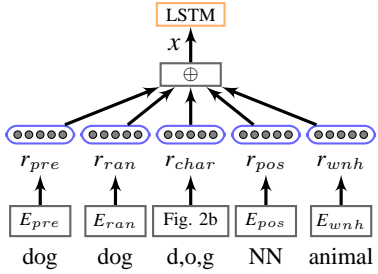
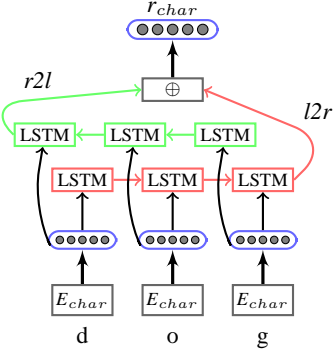

Motivated by prior work \cite{2014:nndep:chen,xu-EtAl:2015:EMNLP2}, other features can also be represented as fixed-length embeddings besides words. We explore five kinds of features in our model, namely pre-trained word features, random word features, character features, POS features and WordNet hypernym features. As shown in Figure \ref{fig:features:concat}, given a token \textquotedblleft{}dog\textquotedblright{}, its pre-trained word, random word, character, POS and WordNet hypernym features are \textquotedblleft{}dog\textquotedblright{}, \textquotedblleft{}dog\textquotedblright{}, \textquotedblleft{}d,o,g\textquotedblright{}, \textquotedblleft{}NN\textquotedblright{} and \textquotedblleft{}animal\textquotedblright{}, respectively. \emph{n}$^{(pre)}$, \emph{n}$^{(ran)}$, \emph{n}$^{(pos)}$ and \emph{n}$^{(wnh)}$-dimensional feature embeddings, namely \emph{r}$_{pre}$, \emph{r}$_{ran}$, \emph{r}$_{pos}$ and \emph{r}$_{wnh}$, are directly taken from their corresponding lookup tables, namely \emph{E}$_{pre}$, \emph{E}$_{ran}$, \emph{E}$_{pos}$ and \emph{E}$_{wnh}$. Since the character number of a word is variable, character features are transformed into a \emph{n}$^{(char)}$-dimensional embedding \emph{r}$_{char}$ using another Bi-LSTM network as shown in Figure \ref{fig:features:char}. \emph{l2r} denotes the last output generated by a LSTM unit computing from left to right, and \emph{r2l} denotes the last output generated by another LSTM unit computing in the reverse direction. The embedding r$_{char}$ of character features is computed by
\begin{equation}
\emph{r}_{char}\ =\ \emph{l2r}\ \oplus\ \emph{r2l}.
\label{eqn:feature:word}
\end{equation}
Finally, we concatenate five kinds of feature embeddings as a composite embedding \emph{x}, given by
\begin{equation}
\emph{x}\ =\ \emph{r}_{pre}\ \oplus\ \emph{r}_{ran}\ \oplus\ \emph{r}_{char}\ \oplus\ \emph{r}_{pos}\ \oplus\ \emph{r}_{wnh}.
\label{eqn:features}
\end{equation}

Pre-trained word features indicate the word features whose embeddings are trained by tools such as word2vec \cite{2013:word2vec:miko} in a great number of external corpora. Most of neural network systems use pre-trained word embeddings to initialize their own word features and tune them in a supervised way during training. Instead, we select pre-trained word embeddings whose domain is consistent with the specific task, and not tune them during training. We believe that pre-trained word embeddings capture global knowledge, which do not need to be adjusted.

Random word features indicate the word features whose embeddings are randomly initialized. By tuning them during training, local knowledge with respect to the specific task can be learned. In our model, both pre-trained and random word features are used, since we believe that they are complementary to each other.

Character features have some distinct characteristics compared with word features. For instance, they can alleviate the out-of-vocabulary problem or capture prefix and suffix information.

POS features are used based on the intuition that the importance of a word for relation classification does not only depend on the word itself, but also its POS tag. For instance, given a \textquotedblleft{}Cause-Effect\textquotedblright{} relation sentence \textquotedblleft{}The \emph{burst} has been caused by water hammer \emph{pressure}.\textquotedblright{}, the verb \textquotedblleft{}caused\textquotedblright{} plays more important role than other words in relation classification. By contrast, the preposition \textquotedblleft{}in\textquotedblright{} is an obvious mark to identify the \textquotedblleft{}Component-Whole\textquotedblright{} relation, given a sentence \textquotedblleft{}The \emph{introduction} in the \emph{book} is a summary of what is in the text.\textquotedblright{}. In this paper, we utilize Stanford CoreNLP toolkit \cite{2014:corenlp:manning} for POS tagging.

WordNet hypernym features come from WordNet \cite{1995:wordnet}, which includes more than 90,000 word senses called synsets. Each noun, verb or adjective synset is associated with one of about 47 broad semantic categories called supersenses (a.k.a., WordNet hypernyms) \cite{socher-EtAl:2012:EMNLP-CoNLL}. For example, given a sentence \textquotedblleft{}My dog ate a bag full of dog treats on Tuesday\textquotedblright{}, its WordNet hypernym annotations will be \textquotedblleft{}My$_{o}$ dog$_{n.animal}$ ate$_{v.consumption}$ a$_{o}$ bag$_{n.artifact}$ full$_{a.all}$ of$_{o}$ dog$_{n.animal}$ treats$_{v.body}$ on$_{o}$ Tuesday$_{n.time}$\textquotedblright{}. \emph{n}, \emph{v}, \emph{a} and \emph{o} indicate \emph{noun}, \emph{verb}, \emph{adjective} and \emph{other}, respectively. WordNet hypernym features are proved to be effective since they reflect word senses, which may be helpful for semantic relation classification \cite{socher-EtAl:2012:EMNLP-CoNLL}. In this paper, we utilize sst-light \cite{ciaramita-altun:2006:EMNLP} for WordNet hypernym tagging.

\section{Experiments} \label{sec:experiment}

\subsection{SemEval-2010 Task 8}

\noindent \textbf{Data and Evaluation Metrics}\\
\indent This dataset \cite{hendrickx-EtAl:2010:SemEval} defines 9 directed relation types between two target entities and one undirected \emph{Other} type when two target entities have none of these relations. We treat each directed relation type as two relation types, so there are totally 19 relation types in our model. The dataset consists of 8,000 training and 2,717 test sentences, and each sentence is annotated with one relation type. Following previous work \cite{hendrickx-EtAl:2010:SemEval,socher-EtAl:2012:EMNLP-CoNLL}, the official macro-averaged F$_1$-score (F$_1$) is used to evaluate performance of different models.
\\\\
\noindent \textbf{Parameter Settings}\\
\indent Parameters are tuned based on the development set, which includes 800 sentences selected from the training set randomly. As it is infeasible to perform full search for all the parameters, some of the values are chosen empirically following prior work \cite{zeng-EtAl:2014:Coling,socher-EtAl:2012:EMNLP-CoNLL,xu-EtAl:2015:EMNLP2}. The initial AdaGrad learning rate $\alpha$ is set as 0.01 and L$_2$ regularization parameter $\beta$ is set as 10$^{-8}$. The dimension of pre-trained word embeddings, \emph{n}$^{(pre)}$  is set as 200. The dimensions of other feature embeddings, namely \emph{n}$^{(ran)}$, \emph{n}$^{(pos)}$, \emph{n}$^{(wnh)}$ and \emph{n}$^{(char)}$, are set as 50. The dimensions of LSTM hidden state (\emph{n}$^{(lstm)}$) and token representation (\emph{n}$^{(h)}$) are set as 200.

The weight matrices \emph{W}, bias vectors \emph{b} and embedding lookup tables \emph{E}$_{ran}$, \emph{E}$_{char}$, \emph{E}$_{pos}$, \emph{E}$_{wnh}$, are randomly initialized in the range (-0.01, 0.01) with a uniform distribution. As for the pre-trained word lookup table \emph{E}$_{pre}$, we train embeddings to initialize it via the snapshot of English Wikipedia\footnote{https://dumps.wikimedia.org/enwiki/} in April, 2016 and word2vec \cite{2013:word2vec:miko} with the skip-gram architecture. The Wikipedia text is preprocessed in the following steps: non-English characters or words are removed; a sentence is removed if it is too short; text is tokenized and all the tokens are transformed into their lowercase forms. Feature embeddings are tuned during training except pre-trained word embeddings.
\\\\
\noindent \textbf{Results}\\
\indent The experimental results on the test set are shown in Table \ref{tbl:result:semeval}. MVRNN \cite{socher-EtAl:2012:EMNLP-CoNLL}, C-RNN \cite{ebrahimi-dou:2015:NAACL-HLT} and DepNN \cite{liu-EtAl:2015:ACL-IJCNLP} are based on RecursiveNNs, but DepNN also combines CNNs to capture features of the shortest dependency paths and further improves the result to 83.6\%. FCM \cite{nips:2014:yumo} achieves a comparable result by decomposing sentences into factors, extracting features and combining them via sum-pooling. CNN-based depLCNN \cite{xu-EtAl:2015:EMNLP1} and RNN-based SDP-LSTM \cite{xu-EtAl:2015:EMNLP2} classify relations using the shortest dependency paths between two entities and obtain similar results. After taking the relation directionality into consideration by a negative sampling strategy, depLCNN achieves state-of-the-art performance (85.6\%). Inspired by \cite{mou-EtAl:2015:EMNLP}, we also experiment with dependency features, and the best result of our model can be 83.1\%. The models mentioned above use structure features, while CNN \cite{zeng-EtAl:2014:Coling} and CR-CNN \cite{dossantos-xiang-zhou:2015:ACL-IJCNLP} only use sequence features such as words and word positions. CR-CNN can achieve 84.1\% in F$_1$ with special treatment for noisy \emph{Other} class, but its F$_1$ is 82.7\% without such special treatment. Our model obtains slightly lower but comparable performance compared with them.

Any kind of models is not absolutely superior to others since they use different features or special treatment. However, the models using structure features usually obtain better performance. This may be because structure features can help removing less relevant noise and providing more compact representations for models. Meanwhile, the shortest dependency paths can take relation directionality into consideration, which may meet the characteristics of this task.

Table \ref{tbl:featcontri:semeval} shows the contributions of different features in our model. By using only pre-trained word features, our model can achieve 78.8\% in F$_1$. WordNet hypernym features are the most effective features, improving F$_1$ from 78.8\% to 79.8\%. Character features are less effective than others, improving F$_1$ by 0.5\%.

\begin{table}[t]
\renewcommand{\arraystretch}{1.3}
\caption{Comparisons with other published results (\%) of neural network models. NER denotes the features of named entity recognition.} \label{tbl:result:semeval}
\begin{center}
\begin{tabular}{|c|p{4.7cm}|c|}
\hline \textbf{Approaches} & \multicolumn{1}{c|}{\textbf{Features}} & \textbf{F$_1$} \\
\hline
MV-RNN & word, POS, NER, WordNet, syntactic & 82.4 \\
C-RNN & word, POS, NER, WordNet, dependency & 82.7\\
FCM & word, NER, depedency & 83.0\\
DepNN & word, NER, depedency & 83.6 \\
depLCNN & word, WordNet, depedency & 83.7\\
SDP-LSTM & word, POS, WordNet, dependency & 83.7 \\
\hline
CNN & word, word position, WordNet & 82.7 \\
CR-CNN & word, word position & 82.7 \\
\hline
Our model & word, char, POS, WordNet & 82.0 \\
Our model & word, char, POS, WordNet, dependency & 83.1 \\
\hline
\end{tabular}
\end{center}
\end{table}

\begin{table}[t]
\renewcommand{\arraystretch}{1.3}
\caption{Feature Contributions (\%) in SemEval-2010 Task 8. Here \textquotedblleft{}+\textquotedblright{} means only one kind of features is added.}
\label{tbl:featcontri:semeval}
\centering
       \begin{tabular}{|c|c|}
       \hline
       \textbf{Features} & \textbf{F$_1$}\\
       \hline
       pretrained word & 78.8\\
       +random word & 79.4\\
       +character & 79.3\\
       +POS & 79.6\\
       +WordNet & 79.8\\
       \hline
       \end{tabular}
\end{table}

\subsection{BioNLP-ST 2016 Task BB3}
Although structure features are useful for relation classification, they are probably domain-dependent. Moreover, there are about 26\% relations between entities that occur in different sentences based on our statistics for BioNLP-ST 2016 Task BB3 \cite{OverviewBB2016}. Structure features are not easy to be directly used since they are designed for using inside one sentence. We experiment on this dataset to prove that our model is still effective even if the problems above exist.
\\\\
\noindent \textbf{Data and Evaluation Metrics}\\
\indent This task includes several subtasks and we focus on the relation classification subtask. The subtask considers one relation type, namely \emph{Lives\_In}, which indicates that bacteria live in a habitat. The dataset consists of 61, 34 and 51 documents for training, development and test, respectively. There are 1080, 730, 1093 entities and 327, 223, 340 relations in the training, development, test sets. We use the official evaluation service\footnote{http://bibliome.jouy.inra.fr/demo/BioNLP-ST-2016-Evaluation/index.html} to evaluate our model. The evaluation metrics are standard precision (P), recall (R) and F$_1$-score (F$_1$).
\\\\
\noindent \textbf{Parameter Settings}\\
\indent Parameters are tuned based on the official development set with 34 documents. The dimensions of pre-trained word embeddings (\emph{n}$^{(pre)}$ ) and random word embeddings (\emph{n}$^{(ran)}$) are set as 200. The dimensions of other feature embeddings, namely \emph{n}$^{(pos)}$, \emph{n}$^{(wnh)}$ and \emph{n}$^{(char)}$, are set as 50. The dimensions of LSTM hidden state (\emph{n}$^{(lstm)}$) and token representation (\emph{n}$^{(h)}$) are set as 200. Other parameter settings are similar to those in the previous task.

\begin{table}[t]
\renewcommand{\arraystretch}{1.3}
\caption{Comparisons with the top 3 results (\%) in the official evaluation. \textquotedblleft{}$\dag$\textquotedblright{} and \textquotedblleft{}$\ddag$\textquotedblright{} denote our model considers relations between entities that occur in the same sentence and two different sentences, respectively.} \label{tbl:result:bb3}
\begin{center}
\begin{tabular}{|c|c|c|c|}
\hline \textbf{Team} & \textbf{F}$_1$ & \textbf{Recall} & \textbf{Precision}  \\
\hline
VERSE & \textbf{55.8} & 61.5 & 51.0 \\
TurkuNLP & 52.1 & 44.8 & \textbf{62.3}\\
LIMSI & 48.5 & \textbf{64.6} & 38.8\\
\hline
Our$^\dag$ & 49.8 & 43.2 & 58.7\\
Our$^\ddag$ & 51.3 & 48.5 & 54.5\\
\hline
\end{tabular}
\end{center}

\end{table}

\begin{table}[t]
\renewcommand{\arraystretch}{1.3}
\caption{Feature Contributions (\%) in BioNLP-ST 2016 Task BB3. Here \textquotedblleft{}+\textquotedblright{} means only one kind of features is added.}
\label{tbl:featcontri:bb3}
\centering
       \begin{tabular}{|c|c|c|c|}
       \hline
       \textbf{Features} & \textbf{F}$_1$ & \textbf{Recall} & \textbf{Precision}  \\
       \hline
       pretrained word & 41.3 & 29.8 & 67.3\\
       +random word & 44.6 & 33.8 & 65.7\\
       +character & 43.9 & 34.4 & 60.9\\
       +POS & 41.9 & 30.1 & 68.9\\
       +WordNet & 44.8 & 34.3 & 64.8\\
       \hline
       \end{tabular}
\end{table}

The weight matrices \emph{W}, bias vectors \emph{b} and embedding lookup tables \emph{E}$_{ran}$, \emph{E}$_{char}$, \emph{E}$_{pos}$, \emph{E}$_{wnh}$, are randomly initialized in the range (-0.01, 0.01). We use biomedical word embeddings \cite{2013:bioemb:Pyysalo} trained from PubMed text to initialize our pre-trained word lookup table \emph{E}$_{pre}$. Feature embeddings are tuned during training except pre-trained word embeddings.
\\\\
\noindent \textbf{Results}\\
\indent The experimental results on the test set are shown in Table \ref{tbl:result:bb3}. VERSE obtains state-of-the-art F$_1$ (55.8\%) in the official evaluation. TurkuNLP and LIMSI achieve the best precision and recall, respectively. When our model considers relations between bacteria/habitat entities that occur in the same sentence, it can obtain better F$_1$ than that of LIMSI. When our model considers relations between bacteria/habitat entities that occur in two continuous sentences, F$_1$ increases from 49.8\% to 51.3\%. If the sentence window is further enlarged, F$_1$ goes down. This may be because most bacteria/habitat entity pairs spanning more than two sentences have no \emph{Lives\_In} relations, the numbers of positive (15\%) and negative (85\%) examples for training the model become very imbalanced.

Feature contributions are shown in Table \ref{tbl:featcontri:bb3}. Our model obtains 41.3\% in F$_1$ using only pre-trained word features. WordNet hypernym features are the most effective features, improving F$_1$ from 41.3\% to 44.8\%. Random word features are more helpful than character features. POS features are less effective than any other kind of features, improving F$_1$ by 0.6\%.

\section{Discussion}

\begin{table}[t]
\renewcommand{\arraystretch}{1.3}
\caption{Context contributions (\%). By default, the \emph{former} and \emph{latter} entity representations are used. The context representations are added, one at a time.}
\label{tbl:context}
\centering
\subfloat[][SemEval-2010 Task 8]{
\centering
       \begin{tabular}{|p{1.4cm}<{\centering}|p{1cm}<{\centering}|}
       \hline
       \textbf{Contexts} & \textbf{F$_1$}\\
       \hline
       \emph{before} & 63.2\\
       \emph{middle} & 81.1\\
       \emph{after} & 60.8\\
       \hline
       \end{tabular}
\label{tbl:context:semeval}
}
\hfil
\subfloat[][BioNLP-ST 2016 Task BB3]{
\centering
       \begin{tabular}{|p{1.3cm}<{\centering}|c|c|c|}
       \hline
       \textbf{Contexts} & \textbf{F}$_1$ & \textbf{Recall} & \textbf{Precision}  \\
       \hline
       \emph{before} & 46.4 & 37.1 & 61.7\\
       \emph{middle} & 47.1 & 38.2 & 61.3\\
       \emph{after} & 45.2 & 36.8 & 58.6\\
       \hline
       \end{tabular}
\label{tbl:context:bb3}
}

\end{table}

\subsection{Which context contributes the most?}
We evaluate contributions of the \emph{before}, \emph{middle} and \emph{after} contexts for relation classification between two target entities. As shown in Table \ref{tbl:context}, the \emph{middle} context plays the most important role. Our model can obtain F$_1$ 81.1\% and 47.1\% using only the \emph{middle} context in SemEval-2010 Task 8 and BioNLP-ST 2016 Task BB3, respectively. The effects of \emph{before} and \emph{after} contexts are almost the same in two datasets, but they are less helpful than the \emph{middle} context. This is consistent with linguistic intuition, since key words or phrases for relation classification are often located in the \emph{middle} context. By contrast, the \emph{before} context often consists of pronouns, articles or modal verbs, and the \emph{after} context often consists of punctuations or complement constituents. More noise in the \emph{before} and \emph{after} contexts lead them to be less helpful for relation classification.

In SemEval-2010 Task 8, the contribution differences between the \emph{middle} and other contexts are more obvious than those in BioNLP-ST 2016 Task BB3. This may be because the \emph{middle} context of a sentence in SemEval-2010 Task 8 is usually much longer than the other two contexts and key words or phrases for relation classification often occur in the \emph{middle} context. By contrast, since the dataset of BioNLP-ST 2016 Task BB3 comes from biomedical publications, there are less key words or phrases to indicate relations obviously and relations are usually implicit in all the contexts.

\subsection{What does the \emph{middle} context capture?}
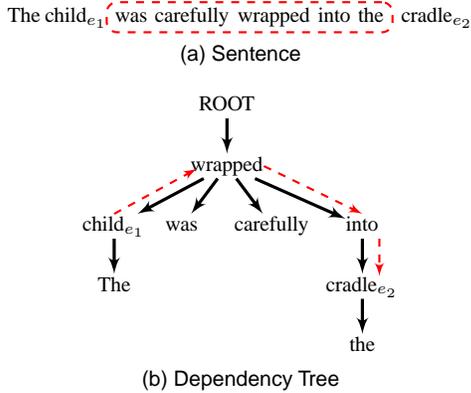
\begin{figure}[t]
\centering

\subfloat[][Sentence]{
\centering
\begin{tikzpicture}[node distance = 2.0, auto]

\node [inner sep=0pt, font=\footnotesize] (middle) {\strut was carefully wrapped into the};
\node [inner sep=0pt, font=\footnotesize, left of=middle, node distance = 2.3cm] (former) {\strut child$_{e_1}$};
\node [inner sep=0pt, font=\footnotesize, left of=former, node distance = 0.7cm] (before) {\strut The};
\node [inner sep=0pt, font=\footnotesize, right of=middle, node distance = 2.5cm] (latter) {\strut cradle$_{e_2}$};
\node [compose5, below of = middle, minimum height = 0.4cm, minimum width = 3.8cm, node distance = 0.0cm]  (comp){};

\end{tikzpicture}
\label{fig:sdpexample:sdp}
}

\hfil

\subfloat[][Dependency Tree]{
\centering
\begin{tikzpicture}[node distance = 2.0, auto]

\node [inner sep=0pt, font=\footnotesize] (x0) {\strut ROOT};

\node [inner sep=0pt, font=\footnotesize, below of=x0, node distance = 0.8cm] (x5) {\strut wrapped};
\path [line] (x0) -- (x5);

\node [inner sep=0pt, font=\footnotesize, below of=x5, node distance = 0.8cm, xshift=-1.5cm] (x2) {\strut child$_{e_1}$};
\node [inner sep=0pt, font=\footnotesize, below of=x5, node distance = 0.8cm, xshift=-0.6cm] (x3) {\strut was};
\node [inner sep=0pt, font=\footnotesize, below of=x5, node distance = 0.8cm, xshift=0.6cm] (x4) {\strut carefully};
\node [inner sep=0pt, font=\footnotesize, below of=x5, node distance = 0.8cm, xshift=1.8cm] (x6) {\strut into};

\path [line] (x5) -- (x2);
\path [line] (x5) -- (x3);
\path [line] (x5) -- (x4);
\path [line] (x5) -- (x6);

\node [inner sep=0pt, font=\footnotesize, below of=x2, node distance = 0.8cm] (x1) {\strut The};
\node [inner sep=0pt, font=\footnotesize, below of=x6, node distance = 0.8cm] (x8) {\strut cradle$_{e_2}$};

\path [line] (x2) -- (x1);
\path [line] (x6) -- (x8);

\node [inner sep=0pt, font=\footnotesize, below of=x8, node distance = 0.8cm] (x7) {\strut the};

\path [line] (x8) -- (x7);

\draw [thick,->,red,-latex',dashed] (x2.north)--+(1.1,0.6);
\draw [thick,->,red,-latex',dashed] (x5.east)--(x6.north);
\draw [thick,->,red,-latex',dashed] (x6.south east)--+(0,-0.5);

\end{tikzpicture}
\label{fig:sdpexample:example}
}

\caption{A sentence and its corresponding dependency tree. Two target entities are \textquotedblleft{}child\textquotedblright{} and \textquotedblleft{}cradle\textquotedblright{}, respectively. Red dashed lines denote the shortest dependency path between target entities.
} \label{fig:sdpexample}
\end{figure}

Prior work \cite{ebrahimi-dou:2015:NAACL-HLT,liu-EtAl:2015:ACL-IJCNLP,xu-EtAl:2015:EMNLP1,xu-EtAl:2015:EMNLP2} has proved that the shortest dependency path (SDP) between two target entities is effective for semantic relation classification, since the words along the SDP concentrate on most relevant information while diminishing less relevant noise. In this subsection, we investigate the relevance between the \emph{middle} context and SDP. A case study is illustrated in Figure \ref{fig:sdpexample}. The \emph{middle} context consists of five words, namely \textquotedblleft{}was carefully wrapped into the\textquotedblright{}. By contrast, the SDP between two target entities in the dependency tree, consists of only two words, namely \textquotedblleft{}wrapped into\textquotedblright{}. The \emph{middle} context captures the information of SDP but also includes some noise.

\begin{figure}[t]
\centering

\subfloat[][SemEval-2010 Task 8]{
\centering
\begin{tikzpicture}[node distance = 2.0, auto]

\node [] (middle_northwest) {};
\node [right of=middle_northwest, node distance = 5.53cm] (middle_northeast) {};
\node [below of=middle_northeast, node distance = 2cm] (middle_southeast) {};
\path [redrect, opacity=0.5] (middle_northwest) rectangle (middle_southeast);

\draw [decorate,decoration={brace,amplitude=10pt}]
(middle_northwest) -- (middle_northeast) node [black,midway,yshift=0.4cm]
{\footnotesize \emph{middle}};

\node [right of=middle_northwest, node distance = 6cm] (sdp_northeast) {};
\node [below of=sdp_northeast, node distance = 2cm] (sdp_southeast) {};
\node [left of=sdp_southeast, node distance = 2.74cm] (sdp_southwest) {};
\path [greenrect, opacity=0.5] (sdp_southwest) rectangle (sdp_northeast);

\draw [decorate,decoration={brace,amplitude=10pt}]
(sdp_southeast) -- (sdp_southwest) node [black,midway,yshift=-0.4cm]
{\footnotesize SDP};

\node [left of=sdp_northeast, node distance = 1.5cm, yshift=-1cm] () {82\%};
\node [left of=sdp_northeast, node distance = 0.2cm, yshift=-1cm] () {18\%};

\end{tikzpicture}
\label{fig:intersect:semeval}
}

\hfil

\subfloat[][BioNLP-ST 2016 Task BB3]{
\centering
\begin{tikzpicture}[node distance = 2.0, auto]

\node [] (middle_northwest) {};
\node [right of=middle_northwest, node distance = 5.01cm] (middle_northeast) {};
\node [below of=middle_northeast, node distance = 2cm] (middle_southeast) {};
\path [redrect, opacity=0.5] (middle_northwest) rectangle (middle_southeast);

\draw [decorate,decoration={brace,amplitude=10pt}]
(middle_northwest) -- (middle_northeast) node [black,midway,yshift=0.4cm]
{\footnotesize \emph{middle}};

\node [right of=middle_northwest, node distance = 6cm] (sdp_northeast) {};
\node [below of=sdp_northeast, node distance = 2cm] (sdp_southeast) {};
\node [left of=sdp_southeast, node distance = 2.51cm] (sdp_southwest) {};
\path [greenrect, opacity=0.5] (sdp_southwest) rectangle (sdp_northeast);

\draw [decorate,decoration={brace,amplitude=10pt}]
(sdp_southeast) -- (sdp_southwest) node [black,midway,yshift=-0.4cm]
{\footnotesize SDP};

\node [left of=sdp_northeast, node distance = 1.7cm, yshift=-1cm] () {61\%};
\node [left of=sdp_northeast, node distance = 0.5cm, yshift=-1cm] () {39\%};

\end{tikzpicture}
\label{fig:intersect:bb3}
}

\caption{An illustration of proportions that the words in the SDPs simultaneously occur in the \emph{middle} contexts or not.
} \label{fig:intersect}
\end{figure}
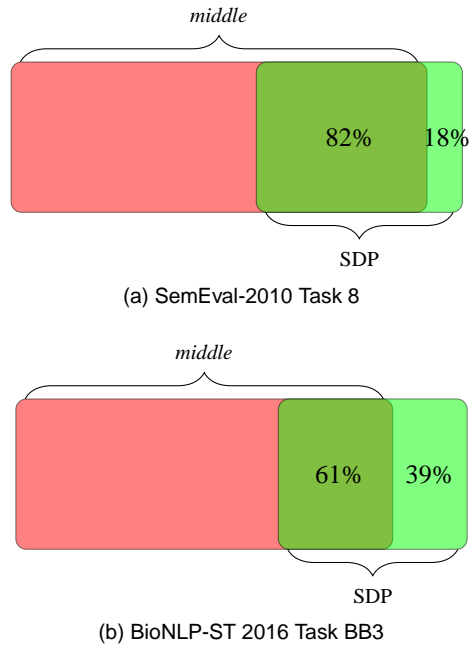

To further prove this, we performs some statistic experiments to count the numbers of words in the \emph{middle} contexts, in the SDPs and occurring in both of them. The experimental steps in SemEval-2010 Task 8 are as follows. First of all, we use Stanford CoreNLP toolkit \cite{2014:corenlp:manning} to perform dependency parsing for all the 8,000 sentences in the training set. Secondly, the SDP between two target entities is built for each sentence. Lastly, we count the numbers of words in the \emph{middle} contexts (26940 words), in the SDPs (13360 words) and occurring in both of them (11054 words). As shown in Figure \ref{fig:intersect:semeval}, about 82\% words in the SDPs occur in the \emph{middle} contexts at the same time.

The experimental steps in BioNLP-ST 2016 Task BB3 are similar. We also use Stanford CoreNLP toolkit \cite{2014:corenlp:manning} to perform dependency parsing for all the 61 documents in the training set. However, there are some differences due to the characteristics of this dataset. Since entities may have more than one words, we use the last words of two target entities to find the SDP in the dependency tree. In addition, only a relation between two target entities that occur in the same sentence, is taken into account, since a dependency tree derives from only one sentence. The numbers of words in the \emph{middle} contexts, in the SDPs and occurring in both of them are 1537, 769 and 466, respectively. As shown in Figure \ref{fig:intersect:bb3}, although the proportion is lower than that in SemEval-2010 Task 8, there are still more than half (61\%) of words in the SDPs occurring in the \emph{middle} contexts at the same time. In this dataset, the \emph{middle} contexts include more words which are not in the SDPs. This may be because the dataset comes from biomedical publications, whose text is often very long with many symbols and numbers.

From the statistic results of our experiments, we believe that the \emph{middle} context captures most of information in the SDP. This suggests that the \emph{middle} context can be used as an approximate replacement of SDP when high-cost dependency parsing is not used.

\section{Conclusion}
We propose a Bi-LSTM-RNN model based on low-cost sequence features to address relation classification. Our motivation is that the relation between two target entities can be represented by the entities and contexts surrounding them. We avoid using structure features to make the model adapt for more domains. Experimental results on two benchmark datasets prove the effectiveness of our model, and its performance gets close to that of state-of-the-art models. By evaluating the contributions of different contexts, we find that the \emph{middle} context plays the most important role in relation classification. Moreover, we also find that the \emph{middle} context can replace the shortest dependency path approximately when dependency parsing is not used. In future work, how to reduce noisy information in contexts is worth studying.

\bibliographystyle{IEEEtran}
\bibliography{IEEEabrv,lstmrnn}
%
%
%

\end{document}